# Architecture of an Ontology-Based Domain-Specific Natural Language Question Answering System

Athira P. M., Sreeja M. and P. C. Reghuraj

Department of Computer Science and Engineering, Government Engineering College, Sreekrishnapuram, Palakkad Kerala, India, 678633

## ABSTRACT

*Question answering (QA) system aims at retrieving precise information from a large collection of documents against a query. This paper describes the architecture of a Natural Language Question Answering (NLQA) system for a specific domain based on the ontological information, a step towards semantic web question answering. The proposed architecture defines four basic modules suitable for enhancing current QA capabilities with the ability of processing complex questions. The first module was the question processing, which analyses and classifies the question and also reformulates the user query. The second module allows the process of retrieving the relevant documents. The next module processes the retrieved documents, and the last module performs the extraction and generation of a response. Natural language processing techniques are used for processing the question and documents and also for answer extraction. Ontology and domain knowledge are used for reformulating queries and identifying the relations. The aim of the system is to generate short and specific answer to the question that is asked in the natural language in a specific domain. We have achieved 94 % accuracy of natural language question answering in our implementation.*

## 1. INTRODUCTION

Question Answering is the process of extracting answers to natural language questions. A QA system takes questions in natural language as input, searches for answers in a set of documents, and extracts and frames concise answers. QA systems provide answers to the natural language questions by considering an archive of documents. Instead of providing the precise answers, in most of the current information retrieval systems the users have to select the required information from a ranked list of documents. Information Extraction (IE) is the name given to any process which selectively structures and combines data which is found, explicitly stated or implied, in one or more texts [5]. After finding the significant documents, the IR system submits those to the user. The scope of the QA has been constrained to domain specific systems, due to the complications in natural language processing (NLP) techniques [4]. Current search engines can return ranked lists of documents, but not the answers to the user queries.

Novice users may lack adequate knowledge in the domain of search, so the query framed by them may not meet the information needs. Moreover, the query that the users often codify captures many documents that are irrelevant, and also fails to find the knowledge or relationships that are hidden in the articles. To overcome this drawback, many systems provide various facilities such as relevance feed-back, with which searchers can find out the documents that are of interest to them. With these questions about the current techniques in mind, a new querying approach can be developed based on domain specific ontologies and some NLP techniques for better results [7]. Also syntactic analysis based on rules and semantic role labeling can be applied to improve both query construction and answer extraction. With this information we will be able to analyze and extract structure and meaning from both questions and candidate sentences, which helps us to identify more relevant and precise answers in a long list of candidate sentences [2].

## 2. LITERATURE SURVEY

There has been an impressive rise in the significance in natural language question answering since the establishment of the Question Answering track in the Text Retrieval Conferences, beginning with TREC-8 in 1999 (Voorhees and Harman, 2000). However, this is not the first time that the QA has been discussed by the NLP researchers. In fact, in 1965 Simmons published a survey article 'Answering English Questions by Computer' and his paper analyses about more than fifteen English language question answering systems implemented in the previous five years [5]. A brief history of QA systems starting from database approaches is briefly described in [4].

Question answering systems have traditionally depended on a variety of lexical resources to bridge the surface differences between questions and potential answers. Syntactic structure matching has been applied to passage retrieval (Cui *et al.*, 2005) and answer extraction (Shen and Klakow, 2006). The significance of semantic roles in answering complex questions was first emphasized by Narayanan and Harabagiu in 2004. <Predicate-argument> structures were identified in their system by consolidating the information on semantic roles from PropBank and FrameNet. But, the history of semantic and thematic role labeling dates back to ancient period. The classical Sanskrit grammar Astadhyayi, created by the Indian grammarian Panini at a time variously estimated at 600 or 300 B.C., includes a sophisticated theory of thematic structure that remains influential till today [2].

Sun *et al.* successfully use semantic relations to match candidate answers. FREyA (Damljanovic *et al.*, 2010) a Feedback Refinement and Extended Vocabulary Aggregation system associates the method of syntactic parsing with ontological information for decreasing the effort of adaptation. In spite of the rule-based systems, their system encodes the knowledge into ontology for a better understanding of the question posed by the user. Then for getting a more definite answer, the syntactic parsing is incorporated.

Our work focuses on the analysis of questions using both syntactic and semantic methods, decomposing a single complex query into a set of less complex queries using an ontology and morphological expansion. Our approach is different from the works in that we use semantic role labeling and domain knowledge using ontology to analyze questions as well as to find answer phrases. Importance is given to both nouns and verbs by extracting the named entities, noun phrases and analyzing then using the Verbnet.

International Journal of Web & Semantic Technology (IJWesT) Vol.4, No.4, October 2013

## 3. QUESTION ANSWERING SYSTEM

Question Answering, the process of extracting answers to natural language questions, is profoundly different from Information Retrieval (IR) or Information Extraction (IE). IR systems present the user with a set of documents that relate to their information need, but do not exactly indicate the correct answer. In IR, the relevant documents are obtained by matching the keywords from user query with a set of index terms from the set of documents. In contrast, IE systems extract the information of interest provided the domain of extraction is well defined. In IE systems, the required information is built around in presumed templates, in the form of slot-fillers.

The QA technology takes both IR and IE a step further, and provides specific and brief answers to open domain questions formulated naturally [9]. Current information retrieval systems allow us to locate documents that might contain the pertinent information, but most of them leave it to the user to extract the useful information from a ranked list [10].

The Question Answering systems based on a repository of documents have three main components. The first is an information retrieval engine that sits on top of the document collection and handles retrieval requests, i.e. a web search engine. The second component is a query interpretation system that deciphers the natural-language questions into keywords or queries for the search engine for fetching the significant documents from the database. That is, the documents that can potentially answer the question [5]. Fine grained information extraction techniques need to be used for pinpointing answers within likely documents. The third component, answer extraction, evaluates these documents and extracts answer snippets from them [8].

The three essential modules in almost all QA systems are question processing (generate a query out of the natural language question), document retrieval (perform document level information retrieval), and answer extraction and formulation (pinpointing answers) [8]. The general architecture of a QA System is shown in Figure 1.

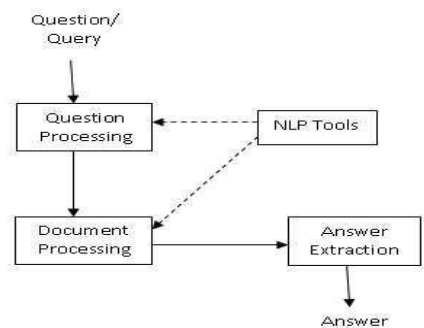

Figure 1: General Architecture of a NLQA System

1. **QUESTION PROCESSING**: The objective of this process is to understand the question posed by the user, for which analytical operations are performed for the representation and classification of the questions.

International Journal of Web & Semantic Technology (IJWesT) Vol.4, No.4, October 2013

2. **DOCUMENT EXTRACTION & PROCESSING:** This module selects a set of relevant documents and extracts a set of paragraphs depending on the focus of the question. The answer is in terms of these paragraphs.

3. **ANSWER PROCESSING:** This module is responsible for selecting the response based on the relevant fragments of the documents. This needs a pre-processing of the information in order to relate the answers with a given question.

Most of the current system uses either syntactic and semantic analysis or ontology processing for answer retrieval. But our system implements the three modules based on a hybrid approach, a step towards natural language question answering in semantic web. The proposed system analyses both the question and answer processing modules using the syntactic and semantic approaches, and also uses the domain ontology for relation extraction and identification. The base ontology is populated dynamically for each document in the collection. And ontology for a specific domain is created as a by-product of the system, which can be used for future analysis and processes.

## 4. THE PROPOSED ARCHITECTURE

The proposed architecture of an ontology-based domain-specific NLQA system is depicted in Figure 2. The model integrates key components such as Natural Language Processing techniques; Conceptual Indexing based Retrieval Mechanism, and Ontology Processing.

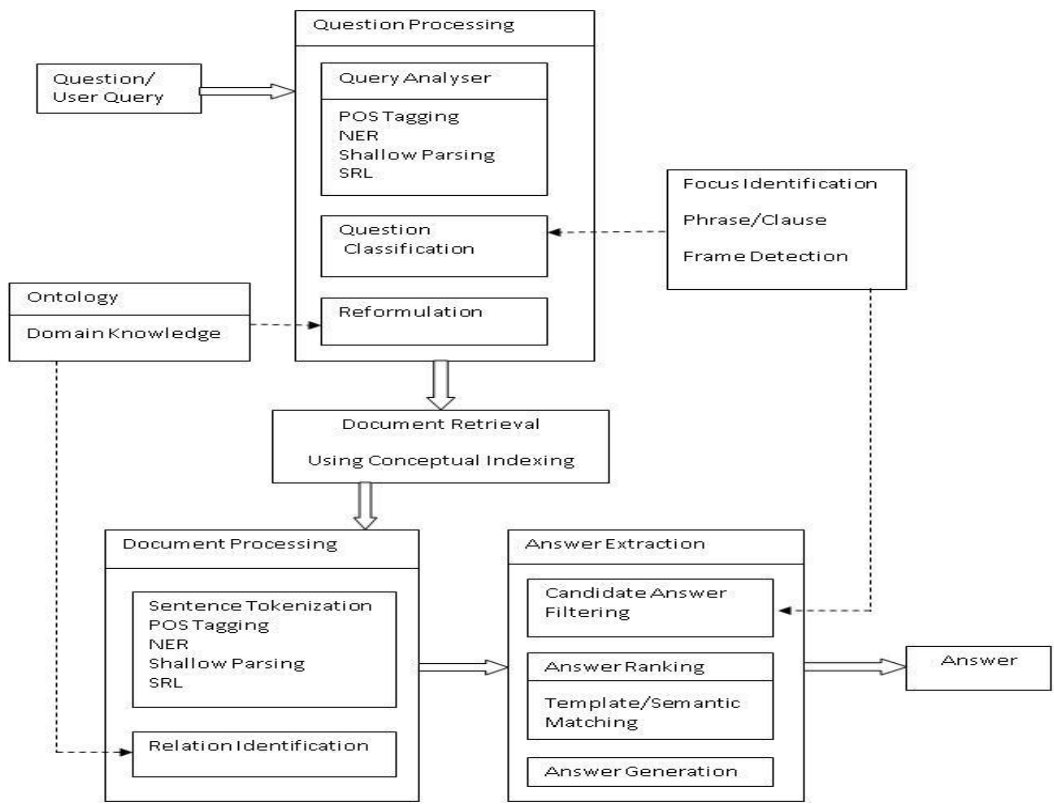

Figure 2: Proposed Architecture of the OD-NLQA System

## 4.1. Question Processing

In the question processing module, with the help of various components, the following actions are performed.

- Analysis of the natural language question
- Question classification
- Reformulation of the user query

### A. Query Analyzer

The natural-language question given by the user is analyzed using various natural language processing techniques.

- Syntactic Analysis – The question is analyzed syntactically using NLP techniques. Part-of-speech tagging and named entity recognition (NER) are performed. Tools such as Python-nltk, OpenNLP, Stanford CoreNLP can be used for this purpose. In the proposed system, we used Stanford CoreNLP tool-kit. The CoreNLP processes the document and creates an XML file as output. Shallow parsing is performed to identify the phrasal chunks. The phrasal chunks can be identified using the Regular-expression chunker and the Conll-2000 trained chunker.

- Semantic Analysis – Semantic role labeling is an important step in this module, which enables to find the dependencies or restriction that, can be imposed, after getting the user query [6]. This greatly eliminates the chances of irrelevant set of answers. Semantic roles are identified using the verbnet frames.

### B. Question Classification

The natural-language question needs to be classified into various sets for extracting more precise sets of answers. The following steps are performed by the proposed system:

- Focus Identification - The objective of this is to identify the category of response that the user is searching for. The question focus can be identified by looking at the question word or a combination of the question word and its succeeding words. Classification of question word to question focus is shown in Table 1. For example, both the question word 'when' or the combination 'what time' indicates a temporal aspect which has to be found in the answer set.

Table 1: Question to Question focus Classification.

| Question Word | Function | Question Focus |
|---|---|---|
| Who | Asking what or which person or people (subject) | PERSON |
| Whom | Asking what or which person or people (object) | PERSON |
| Whose | Asking about ownership | PERSON |
| Where | Asking in or at what place or position | LOCATION |

| When | Asking for time or date | DURATION or DATE |
|---|---|---|
| How + Adj/Adv | Asking about extend or degree | |
| How far | Asking about distance | NUMBER |
| How long | Asking about length (time) | DURATION |
| How many | Asking about quantity (countable) | NUMBER |
| How much | Asking about quantity (uncountable) | METRICS |

- Phrase or Clause Detection - The phrases and clauses contains the information relevant to the expected answer and the irrelevant set can be easily eliminated. The question phrase can be determined using shallow parsing or chunking. The NP chunk identifies the clauses that are to be looked upon in the documents for obtaining the answer set.

- Frame Detection – The semantic roles are identified and mapped to a semantic frame for better retrieval. For example: The event E "Who gave a balloon to the kid?" has the roles "AGENT verb/give THEME to RECIPIENT", the semantic frame is identified as "has_possession(start(E), Agent, Theme ) has_possession(end(E), Recipient, Theme) transfer(during(E), Theme)". The frames can also be used for ranking of the retrieved answer set.

**C. Query Reformulation**

The user queries may be reformulated by adding domain knowledge and ontological information.
- Ontology - Ontology is defined in the basic terms and relations comprising the vocabulary of a specific area, as well as rules for combining these terms and relations to define extensions vocabularies [1]. The base ontology is created for the specific domain by incorporating the classes and object properties. The domain, range, and restrictions on the classes are also specified.

## 4.2. Document Retrieval

This module selects a set of relevant documents from a domain specific repository. Conceptual indexing is used for the retrieval process since the key word based indexing ignores the semantic content of the document collection [5]. Both the documents and queries can be mapped into concepts and these concepts are used as a conceptual indexing space for identifying and extracting documents.

## 4.3 Document Processing

The retrieved documents are processed for extracting candidate answer set. This module is responsible for selecting the response based on the relevant fragments of the documents.

- Syntactic Analysis – The documents analyzed syntactically using the NLP techniques such as part-of-speech tagging and named-entity recognition. In the syntactic analysis, firstly the documents are tokenized into set of sentences. Using the Stanford CoreNLP, the POS tagging and NER is performed. Shallow parsing is performed to identify the phrasal chunks. The chunks identified in the question analysis module are matched with those identified in the document and relevant sentences are retrieved.

- Semantic Analysis – Shallow parsing can be performed for finding the semantic phrases or clauses. The semantic roles are identified and mapped to semantic frames. The sentences whose semantic frames map exactly to the semantic frames of the question are also extracted.

- Relation Identification - The base ontology is populated with the domain knowledge incrementally as we go through different set of documents. By this method a valid knowledge on any specialized discipline can be incorporated to the system. The relations among different concepts are identified using the domain knowledge and the ontological information obtained.

## 4.4 Answer Extraction

The filtering of candidate answer set and answer generation is performed. The user is supplied with a set of short and specific answers ranked according to their relevance. The different stages are:

- Filtering – The extracted sentences are filtered and the candidate answer set is produced. This is done by incorporating the information obtained from the question classification and document processing modules. The identified focus and frames are matched to get the candidate set.

- Answer Ranking – The answer set is ranked based on the semantic similarity. Simple template matching is not adopted since it neglects the semantic content and domain knowledge. Answers are ranked based on the similarity between the question frame and the answer frame. Example: The event E "John gave a balloon to the kid." has the roles "AGENT verb/give THEME to RECIPIENT, the semantic frame is identified as "has_possession(start(E), Agent, Theme ) has_possession(end(E), Recipient, Theme) transfer(during(E), Theme)" matches exactly with the question frame.

- Answer Generation – From the answer set, specific answers have to be generated in case the direct answers are not available. Hidden relations can be identified from the domain knowledge gathered from the ontology. Concept of natural language generation can also be utilized for this purpose.

## 5. EVALUATION AND TESTING

The TREC QA test collections contain newswire articles and the accompanying queries cover a wide variety of topics [11]. QA evaluation process simply access these TREC collections for testing the efficiency of the systems. Simply applying the open domain QA evaluation paradigm to a restricted domain poses several problems. So a different method for evaluation process is used in this system. A random set of documents are collected over a specific domain. Relevant, correct and complete answers are derived for a set of question from some arbitrarily chosen unbiased users, and this set is used for the testing purposes [4]. The questions are tested for their:

- Correctness - The answer should be factually correct
- Relevance - The answer should be a response to the question
- Completeness - The answer should complete, i.e. a partial answer should not get full credit.

The proposed system tries to find precise answers to factual questions and explanative answers are not provided. For multiple answers, ranking is provided based on the semantic matching. Direct answers are generated using natural language generation techniques form the candidate set of answers.

The system is tested for efficiency using the notion of recall. Recall for a question answering system is defined as the "ratio of number of correct answers to the total number of questions given [5]." Answer precision may be subjective, but we have tried to make it as objective as possible.

The system is tested in a domain of short stories and the results are presented in Table 2 and Table 3. For the system testing, the document set (collection of short stories) is retrieved arbitrarily over the web. For the testing we use the question track consisting of 100 questions of varying type complexity and difficulty. It has to be noted that *precise* answer here indicates the one-word answer generated as response to a factual question, but for a question answering system, the answer can be generated as a single sentence and is indicated as retrieved sentences containing *precise* answers.

Table 2: % Recall with retrieved *precise* answers.

| No. of documents | Total size | No. of questions | No. of correct answers | % Recall |
|---|---|---|---|---|
| 20 | 478KB | 50 | 41 | 82 |
| 50 | 1.2MB | 120 | 97 | 80.8 |

Table 3: % Recall considering retrieved sentences containing *precise* answers.

| No. of documents | Total size | No. of questions | No. of correct answers | % Recall |
|---|---|---|---|---|
| 20 | 478KB | 50 | 47 | 94 |
| 50 | 1.2MB | 120 | 112 | 93.3 |

## 6. CONCLUSION AND FUTURE WORK

We have presented an architecture of ontology-based domain-specific natural language question answering that applies semantics and domain knowledge to improve both query construction and answer extraction. The system presented in this paper is a step towards the ultimate goal of using the web as a comprehensive, self-updating knowledge repository, which can be automatically mined to answer a wide range of questions with much less effort than is required by todays search engines. The experiments show that our system is able to filter semantically matching sentences and their relations effectively and therefore, rank the correct answers higher in the result list.

We intend to extend the coverage of the system to all possible question types i.e. move from the factual to more complex forms of question, including lists, summarization of contradictory information, and explanations, including answers to how or why questions, and eventually, what if questions. Generating short, coherent and precise answers will be a major research area and will rely massively on progress in information extraction and text summarization. Also new research should be done to gather more information in various levels of understanding, effectiveness and situations.